# uTHCD: A New Benchmarking for Tamil Handwritten OCR

*Noushath Shaffi and Faizal Hajamohidden*
Dept of Information Technology
University of Technology and Applied Science, Sultanate of Oman.
Email: {noushath.soh, faizalh.soh}@cas.edu.om

**Abstract:**

Handwritten character recognition is a challenging research in the field of document image analysis over many decades due to numerous reasons such as large variation of writing styles, inherent noise in the data, expansive range of applications it offers, non-availability of benchmark databases etc. There has been considerable work reported in the literature about the creation of the database for several Indic scripts but the Dravidian Tamil script is still in its infancy as it has been reported so far only in one database [5]. In this paper, we present the work done in the creation of an exhaustive and large unconstrained Tamil handwritten character database (uTHCD). This database consists of around 91000 samples with nearly 600 samples in each of 156 character classes. The database is a unified collection of both online and offline samples. Offline samples were collected by asking volunteers to write their samples on a form inside a specified grid. For online samples, we made the volunteers write in a similar grid on an electronic form using a digital writing pad. The samples thus collected encompass a vast variety of writing styles, inherent distortions that arise from offline scanning process such as stroke discontinuity, variable thickness of stroke, distortion etc. Algorithms which are resilient to such data can be practically deployed for real time applications. The samples were generated from around 650 native Tamil volunteers including school going kids, homemakers, university students and faculty. The isolated character database will be made publicly available both as raw images as well as HDF compressed file suitable for direct implementation. With this database, we expect to set a new benchmark in Tamil handwritten character recognition research and serve as a launchpad for many avenues in the document image analysis domain. The paper also presents an ideal experimental set-up carried out using the database on a convolutional neural networks (CNN) architecture with a baseline accuracy of 88% on the test data. Furthermore, experiments were conducted to overcome the data overfitting by the model.

The ***Tamils*** or ***Tamilians*** is one of the world's oldest surviving ethnolinguistic groups with a demographic population currently estimated to be around 76 millions with history of this language dating back over 2000 years [1]. Out of 22 official languages of India, it is the only language that has been considered as an official language outside of India as in Srilanka and Singapore [1]. In addition, scripts of several minority Indian languages such as Baduga, Irula, Saurashtra and Paniya are written in Tamil although its spoken version is completely different from that of Tamil [2].

Despite the fact that a sizable population around the globe uses Tamil language as one of the official or as borrowed scripting languages, research in Tamil handwriting recognition in several application domains has not reached its complete maturity. Tamil Handwritten character recognition is one such challenging research topic that has existed for close to 4 decades now [3] and continues to offer a plethora of challenges which keeps the research community active even till date [4]. Although the OCR of printed documents has been considered as a problem solved par human accuracy or even better, the challenge in Tamil Handwritten OCR research is mainly attributed to complex character formation which results in indefinably large writing styles even within the writing of a single user.

The research in this field is equally instigated by several application oriented domains such as document structure analysis and segmentation, writer identification in forensic intelligence, digitization of legal and legacy documents. Since Tamil language is one of the official languages in India ,Srilanka and Singapore, availability of a robust Tamil handwritten OCR would pave way for automation of a large swath of official documents.

Extensive work relating to the Indic script has been reported in this field of handwriting recognition which is evident through numerous publications [5]. This was mainly possible due to the availability of standardized databases of many Indic scripts [7, 8, 9, 10, 11, 12, 13, 14, 15, 16, 17] .There are about 19 standardized handwritten datasets available for Indic scripts such as Devanagari, Bengali, Telugu, Oriya, Tamil etc [5].

Out of 19 databases, as of writing this paper, we found only a single standardized database [5] that is available for the Tamil script developed by HP Labs, India [6, 7]. This database is publicly available and was created using online pen coordinates through the application of simple piecewise interpolation process. Most of the work reported in Tamil handwriting recognition has relied largely on this database. However, for more objective study and to evaluate the true efficacy of state-of-the art algorithms, having multiple standardized publicly available datasets exhibiting a variety of inherent writing variations can only add more value to the resulting experimental analysis.

Unlike the HPL database, the uTHCD database is a unified collection of both offline and online samples. While online samples capture successfully different writing styles, salient features of offline samples pose even more challenges. Specifically sharp boundaries, discontinuity of stroke occurring as a result scanning and several subsequent preprocessing, variable thickness of character stroke due to the usage of different writing tools such as pen, pencil, sketch pens, gel pens etc add more variation to the database. Offline counterpart samples generated through online data as in the HPL database do not represent these variations. Moreover, capturing these variations in the database will prove more effective in the development of robust algorithms especially when it comes to processing offline documents such as legal and legacy documents, handwritten form conversions etc.

The main motivations behind carrying out this work are:

a) Non availability of multiple databases that can pose diverse challenges that are inherent in a typical offline handwriting OCR.
b) Non existence of a standardized handwriting database that can capture variations that emerges from an offline processing of documents.
c) Evaluate objectively several classical machine learning and state-of-the-art deep learning algorithms using the uTHCD database and benchmark the results. This will act as a catalyst in advancing research in Tamil handwriting recognition.

The prominent contribution of this work can be outlined as below:

a) A succinct presentation of the complete process involved in the creation of a comprehensive unconstrained Tamil Handwritten character samples.
b) Establishing a baseline accuracy using CNN and setting a new benchmark for this database.

This work is an extension of the previously published paper where we had reported the preliminary work [18] done in the process of creation of the uTHCD database.

The remainder of this paper is organized as follows: Section 2 describes the details about the Tamil script, overview of existing Tamil handwritten databases and works reported using them. Section-3 presents the systematized process with appropriate signposting involved in the creation of this database such as data collection, preprocessing, verification and ultimate creation of the database suitable for the experimentation process. In section-4, we set the new benchmark for this database through the application of standard CNN algorithms. The final conclusion and future avenues based on this work are presented in section-5.

## 2. Background Study:

**2.1 : Tamil Script:** The Tamil script contains 12 vowels, 18 consonants and one special character ஃ (known as Ayudha Ezhuthu). Additional five consonants known as Grantha Letters are borrowed from Sanskrit and English to represent words/syllables of north Indian and English origin [1]. Hence the script contains 36 unique basic letters [12 vowels + 18 consonants + 1 Ayudha Ezhuthu + 5 Granthas].These basic characters are shown in Fig.1**.**

| | | | | | | | | | | | | | | | | | |
|---|---|---|---|---|---|---|---|---|---|---|---|---|---|---|---|---|---|
| அ | ஆ | இ | ஈ | உ | ஊ | எ | ஏ | ஐ | ஒ | ஓ | ஔ | ஃ | ஜ | ஷ | ஹ | ஸ | க்ஷ |
| a | ä | i | ī | u | ü | e | é | ai | o | ò | au | aytam | j | ṣ | h | s | kṣ |
| க | ச | ங | ஞ | ட | ண | த | ந | ப | ம | ய | ர | ல | ள | ற | வ | ழ | ன |
| k | c | ṅ | ṅ | ṫ | ṇ | t | n | p | m | y | r | l | ḷ | r | v | ẓ | ṇ |

Fig.1 Vowels, Consonants, Ayutha Ezhuthu and Grantha letters of Tamil Script

The combination of 12 vowels and 18 consonants gives rise to formation of 216 compound characters resulting in a total of 247 characters [216 + 12+18+1]. Example formation of compound characters for a consonant is shown in the figure Fig.2.

| | | | | | | | | | | | |
|---|---|---|---|---|---|---|---|---|---|---|---|
| க் + அ | க் + ஆ | க் + இ | க் + ஈ | க் + உ | க் + ஊ | க் + எ | க் + ஏ | க் + ஐ | க் + ஒ | க் + ஓ | க் + ஔ |
| க | கா | கி | கீ | கு | கூ | கெ | கே | கை | கொ | கோ | கௌ |

Fig.2 Compound Character Formation For a Consonant With Set of Vowels

Furthermore, the 5 grantha letters combine with 12 vowels in a similar fashion to form additional 60 compound letters leading to a formation of total 312 characters [247 + 5 Granthas + 60 Compound Granthas]. The entire character set comprising of 312 characters of the script is shown in Figure 3. However, these 312 characters can be represented uniquely with 156 distinct characters (shaded grey). This is because, as depicted in Figure 4, two different syllables of the word [Ko-vai] can be represented using 5 distinct characters (numbers in the figure indicate the underlying class number).

| ஆய்த எழுத்து | | Vowels (உயிர் எழுத்துக்கள்) | | | | | | | | | | | |
|---|---|---|---|---|---|---|---|---|---|---|---|---|---|
| ஃ | | அ | ஆ | இ | ஈ | உ | ஊ | எ | ஏ | ஐ | ஒ | ஓ | ஔ |
| Consonants (மெய் எழுத்துக்கள்) | க் | க | கா | கி | கீ | கு | கூ | கெ | கே | கை | கொ | கோ | கௌ |
| | ங் | ங | ஙா | ஙி | ஙீ | ஙு | ஙூ | ஙெ | ஙே | ஙை | ஙொ | ஙோ | ஙௌ |
| | ச் | ச | சா | சி | சீ | சு | சூ | செ | சே | சை | சொ | சோ | சௌ |
| | ஞ் | ஞ | ஞா | ஞி | ஞீ | ஞு | ஞூ | ஞெ | ஞே | ஞை | ஞொ | ஞோ | ஞௌ |
| | ட் | ட | டா | டி | டீ | டு | டூ | டெ | டே | டை | டொ | டோ | டௌ |
| | ண் | ண | ணா | ணி | ணீ | ணு | ணூ | ணெ | ணே | ணை | ணொ | ணோ | ணௌ |
| | த் | த | தா | தி | தீ | து | தூ | தெ | தே | தை | தொ | தோ | தௌ |
| | ந் | ந | நா | நி | நீ | நு | நூ | நெ | நே | நை | நொ | நோ | நௌ |
| | ப் | ப | பா | பி | பீ | பு | பூ | பெ | பே | பை | பொ | போ | பௌ |
| | ம் | ம | மா | மி | மீ | மு | மூ | மெ | மே | மை | மொ | மோ | மௌ |
| | ய் | ய | யா | யி | யீ | யு | யூ | யெ | யே | யை | யொ | யோ | யௌ |
| | ர் | ர | ரா | ரி | ரீ | ரு | ரூ | ரெ | ரே | ரை | ரொ | ரோ | ரௌ |
| | ல் | ல | லா | லி | லீ | லு | லூ | லெ | லே | லை | லொ | லோ | லௌ |
| | வ் | வ | வா | வி | வீ | வு | வூ | வெ | வே | வை | வொ | வோ | வௌ |
| | ழ் | ழ | ழா | ழி | ழீ | ழு | ழூ | ழெ | ழே | ழை | ழொ | ழோ | ழௌ |
| | ள் | ள | ளா | ளி | ளீ | ளு | ளூ | ளெ | ளே | ளை | ளொ | ளோ | ளௌ |
| | ற் | ற | றா | றி | றீ | று | றூ | றெ | றே | றை | றொ | றோ | றௌ |
| | ன் | ன | னா | னி | னீ | னு | னூ | னெ | னே | னை | னொ | னோ | னௌ |
| Grantha (வடமொழி எழுத்துக்கள்) | ஜ் | ஜ | ஜா | ஜி | ஜீ | ஜு | ஜூ | ஜெ | ஜே | ஜை | ஜொ | ஜோ | ஜௌ |
| | ஷ் | ஷ | ஷா | ஷி | ஷீ | ஷு | ஷூ | ஷெ | ஷே | ஷை | ஷொ | ஷோ | ஷௌ |
| | ஸ் | ஸ | ஸா | ஸி | ஸீ | ஸு | ஸூ | ஸெ | ஸே | ஸை | ஸொ | ஸோ | ஸௌ |
| | ஹ் | ஹ | ஹா | ஹி | ஹீ | ஹு | ஹூ | ஹெ | ஹே | ஹை | ஹொ | ஹோ | ஹௌ |
| | க்ஷ் | க்ஷ | க்ஷா | க்ஷி | க்ஷீ | க்ஷு | க்ஷூ | க்ஷெ | க்ஷே | க்ஷை | க்ஷொ | க்ஷோ | க்ஷௌ |
| | | | | | | | ஸ்ரீ | ெ◌ | ே◌ | ை◌ | ◌ா | | |

Fig. 3 Complete Character Set of Tamil Script

**2.2: Existing Tamil Handwritten Character Database:** Existence of a comprehensive database plays a vital role in the advancement of any research topic. It will facilitate researchers to rapidly evaluate new models and also in the development of robust algorithms especially when the underlying database is unconstrained and comprehensive. Availability of a comprehensive database can also help to objectively evaluate true efficacy/limitations of state-of-the-art algorithms. Developing a robust Tamil OCR system necessitates a large and comprehensive database. As mentioned earlier, there is only a single standardized database developed by HP labs India, available for Tamil OCR domain [5].

**2.2.1: HPL India database:** This database (HPL-iso-tamil-char) contains varying number of isolated character samples in each of 156 distinct character classes. Most of the character classes have around 500 samples while few classes have around 270 samples. Maximum number of samples in a class is 527 while minimum is 271. The database consists of two separate repositories for training and testing. There are around 50691 and 26926 samples available respectively for training and testing purposes. The database contains 77617 samples in total from all classes.

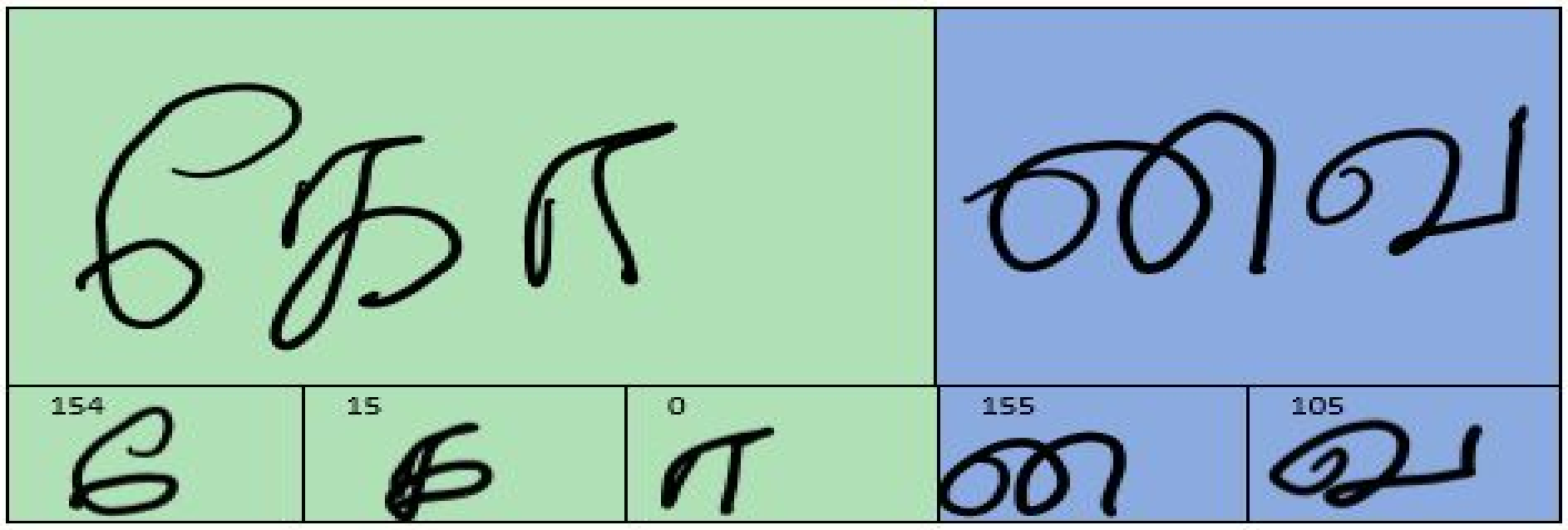

**Fig.4 Two syllables (not among 156 classes) represented by 5 distinct characters of the database.**

The samples were written by native Tamil writers and collected using HP Tablet PCs. As mentioned earlier, images of samples were created through an interpolation process using online pen coordinates with a constant thickening factor and are available for download in a bi-level TIFF format. More details of this database can be found in the lipitoolkit website [6]. This database is the de-facto standard among the research community for Tamil OCR and one can easily find numerous quality papers in the literature that used this database [25, 26, 27, 28, 29]. Very recently, Kavitha et al have used this database to test the efficacy of 9-layer CNN architecture and the proposed architecture achieved a training accuracy of 95.16% and test accuracy of 97.7% [25]. However, this database has following limitations:

1. Imbalance in the dataset - There are several classes that are underrepresented in this collection. The algorithm using this database learns to predict correctly the dominant classes more often than the underrepresented class. As a result of this, algorithms using this collection will have to use different evaluation metrics such as F1 score per class, recall per class, precision per class other than just accuracy. Another option would be to resample the dataset so that the data is balanced either by augmentation or undersampling (throw away examples of dominant classes). Only then the model developed using this database can be suitable for real world applications.

2. All samples are collected by digital means. As mentioned earlier, such samples are not true representatives of samples that are often seen in offline OCR applications.

**2.2.2: Ad-hoc databases:** Going through literature in this field, one can easily find out that there have been many works that present Tamil handwritten OCR by making use of databases that are created in house [19, **20**, 21, 22, 23]. Many works apparently collected very limited data [19, 20]. Kowsalya et al in [20] have proposed an algorithm for tamil OCR by making use of elephant-herd optimization. For experimental purposes, the authors have made use of limited samples collected from different native Tamil speakers. Jose et al in [21] have collected samples from 100 different users with each class having 100 samples. There is no mention of how many classes are considered for data collection and experimentation. Shanti et al in [22] have trained the system with 35441 characters belonging to 106 tamil characters. The database was created through the contribution of 117 volunteers. The test data had 6048 samples belonging to only 34 character classes. Vinotheni et al in [23], have created a database comprising 54600 samples with 350 samples in each of 156 classes. This work does not indicate the train-test split.

It can be ascertained that ad-hoc databases suffer from several drawbacks such as instance in-house creation, publicly not available, incomprehensiveness and insufficient number of samples. This will impede the research community in carrying out comparative analysis of algorithms proposed in the literature. Atleast, if these databases are made publicly available, the research community will have choices and can contribute to easy prototyping and evaluation of new Tamil OCR models.

Based on this survey of literature, following main conclusions can be drawn:

1) Unlike any other languages such as English, or even certain Indic scripts such as Devanagari, Bangla, Oriya etc, there are no multiple standardized publicly available databases for Handwritten Tamil OCR process which could involve diversity of writing styles by many people.
2) Majority of the standard papers deal with HPL-ISO-Tamil-Char dataset. While this database definitely helped to trigger a sizable amount of research in offline Tamil handwriting recognition, it has few limitations as well such as insufficient samples in certain classes and samples not being true representative of offline scanned documents.
3) Various other methods proposed by different authors in the literature are giving good performance but on data prepared in an ad-hoc way. Such an algorithm cannot be reliably deployed for practical applications.

Development of an exhaustive database involving handwriting from many Tamil writers and testing the efficacy of a computational algorithm (CNN) on this exhaustive database is proposed in this study. Our database has two prominent merits as mentioned below and can impact positively the research community.

a) The uHTCD database has 90950 samples with approximately 600 samples from each of 156 classes with provision to enhance more in the coming years. This is approximately 13000 samples more compared to the existing standard HPL-ISO-Tamil-Char database.
b) The database is a unified collection of offline and online samples. In addition to capturing the variety of writing styles, models created using the uHTCD database will capture the inherent characteristics of scanned documents thereby robust performance is expected for automatic form processing.

The uTHCD database will be the second public repository for Tamil OCR purposes. The research community can evaluate the algorithms on multiple databases instead of relying on a single database that contains only online samples. The table.1 provides a quick comparison between these two repositories.

**Table1 Comparison of HPL-ISO-TAMIL-CHAR and uTHCD Databases**

| **Comparison Factors** | **HPL-ISO-TAMIL-CHAR** | **uTHCD** |
|---|---|---|
| Train set samples | 50691 | 62870 |
| Test set samples | 26926 | 28080 |
| Offline samples available | NO | YES |
| Possible future expansion | NO | YES |
| Publicly Available Hierarchical Data Format (HDF) | NO | YES |
| Uniform distribution of samples in each class | NO | YES |

## 3. Database Work:

The database creation is a laborious process which involves many tasks such as sample collection, data extraction, carrying out necessary postprocessing, verification and finally making it ready for experimentation by associating it with ground truth (GT). The overall process involved is as shown in **Fig.5** . All these tasks have to be carried out with utmost care as any mistake would eventually affect the benchmarking process. This section presents all steps involved in the creation of this database.

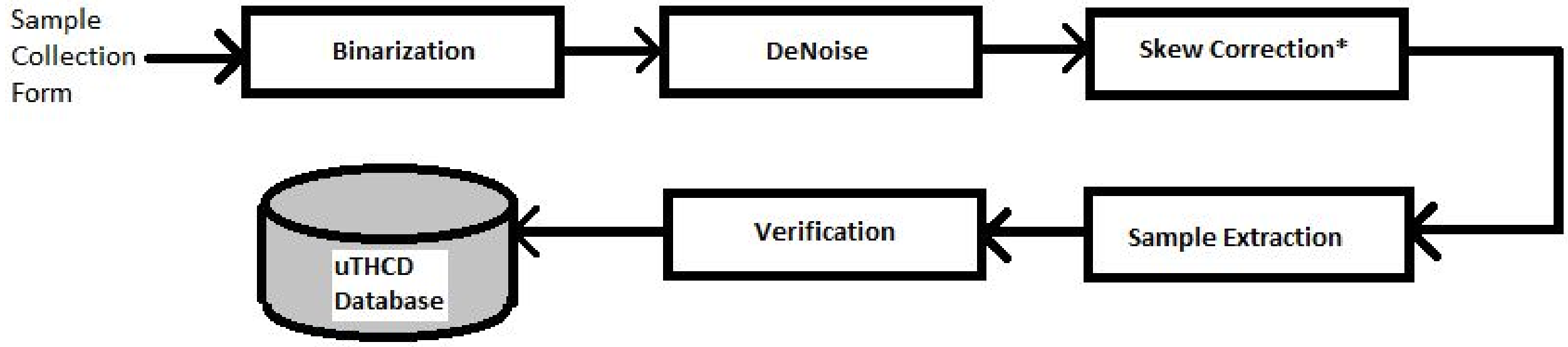

Fig.5 The Important Steps in the Creation of uTHCD Database

### 3.1 Sample Collection

The database is a unified collection of samples collected from offline and online modes. The overall process involved in both are the same except that the online mode was relatively straightforward as it did not necessitate preprocessing the document through skew estimation step.

The data collection process starts with a A4 sized paper where we placed a 10x8 grid asking the volunteers to write the predefined Tamil characters in specific cells. There are two such data collection forms wherein one complete set of samples pertaining to 156 classes would be collected. Each volunteer was requested to fill only one form as casting complete set would cause possible lethargy and may lead to unnecessary processing of samples that are not true representative of the underlying character class. In this way, a complete set of distinct tamil characters could be obtained by two volunteers. A sample form used in this data collection process is shown in **Fig.6.**

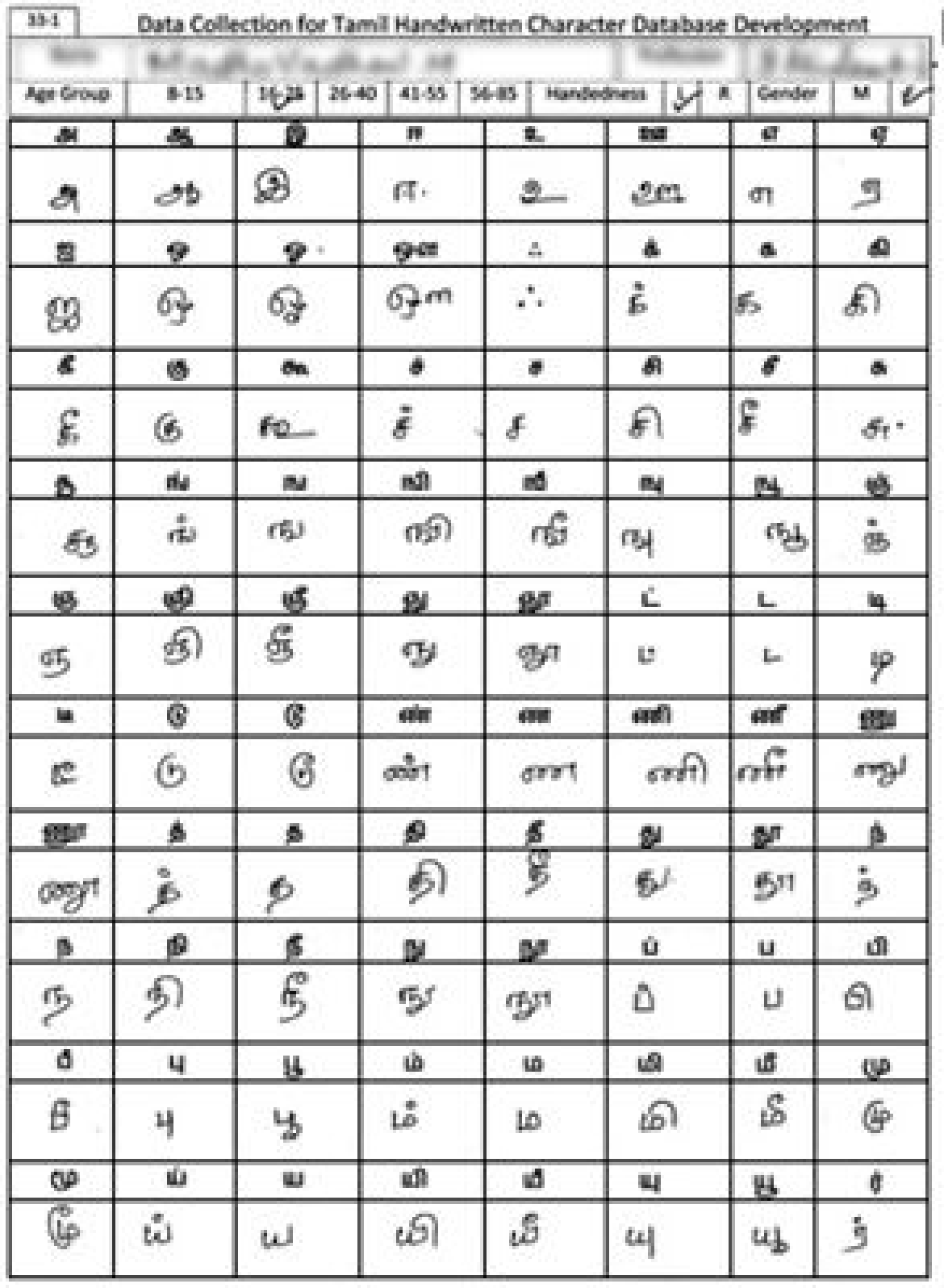

Fig.6 Sample Collection Form

Some quality checks we ensured during this process are:

1) The volunteer was asked to provide the sample within the cells in the grid so as not to overlap with characters in adjacent cells. Negligible amount of samples breached this but it was easy for us to discard such samples during the verification process.
2) Each volunteer must have studied Tamil language formally as primary/secondary language during their school. This ensures that writing style captured in the sample would not drift much from the normal writing style of the Tamil script.

The most time consuming part of data collection was in the offline mode as it involved a waiting period where we could not reliably receive all the forms from volunteers couriered to different parts of the world. We ended up collecting samples from around 400 volunteers over a period of 3 years. This resulted in around 200 sets of distinct offline character samples collected in the offline mode. The offline samples constitute around 30% in the database.

A similar high-resolution grid was designed for an online sample collection process. Each volunteer was asked to write in the predefined space in a manner similar to offline mode. The samples were collected using a digital writing pad. This whole process involved contributions from around 450 volunteers over a period of 2 years. The data collection happened among students and staff belonging to several schools, universities, and at community places where there are sizable native Tamil speakers. The samples thus collected in offline and online mode were curated to contain approximately 600 samples from each class with online samples constituting around 70% of the database.

### 3.2: Preprocessing and Sample Extraction

The offline data collection forms are first digitized at 300 DPI through a HP flatbed multifunctional scanner. The scanned image was then assigned a unique 3 digit number between 001 to 600 indicating the volunteer code. After digitalizing the form, individual character samples were extracted using Python modules. At first, the scanned image was binarized using simple global thresholding. Subsequently median filtering and size based connected component empirical analysis was carried out to get rid of isolated pixels and irrelevant clusters of pixels arising due to various reasons such as dust in the scanner bed, liquid spillage, smudging, etc. The scanned image was then skew corrected using the ePCP algorithm [24]. We exercised extreme caution while feeding the page during the scanning process so as to ensure minimal skew. This helped us to keep the ePCP profiling in the -5 to +5 degrees with increment of 0.1 for each step. The same step was repeated for online samples too except that they were not subjected to skew correction step for obvious reasons.

As both offline and online samples were collected by asking volunteers to write within the grid, we used starting coordinates of the grid and with appropriate height and width increments resulted in accurate extraction of character samples. Samples were aggregated in one of the 156 classes based on their location in the grid. In this way, each class contains samples from both online and offline modes collected from all volunteers. Each sample will be saved using the filename convention: xxxx_yyy where yyy represents the underlying class number and xxxx corresponds to a 4-digit volunteer number starting serially from 0001 to 0600 (additional MSD digit is to accommodate future expansion). This way yyy will also serve as ground truth for the corresponding sample and assist in the GT verification later on during the final verification phase. For instance, samples in class-1 will be named 0001_001, 0002_001, 0003_001, etc. To further distinguish between offline and online samples, letter ‘s’ will be affixed in the format as xxxx**s**_yyyy**.**

Some sample offline and online samples thus extracted are shown in Fig. 7(a) and Fig.7(b) respectively. As mentioned earlier, it is worth noticing that offline samples capture useful

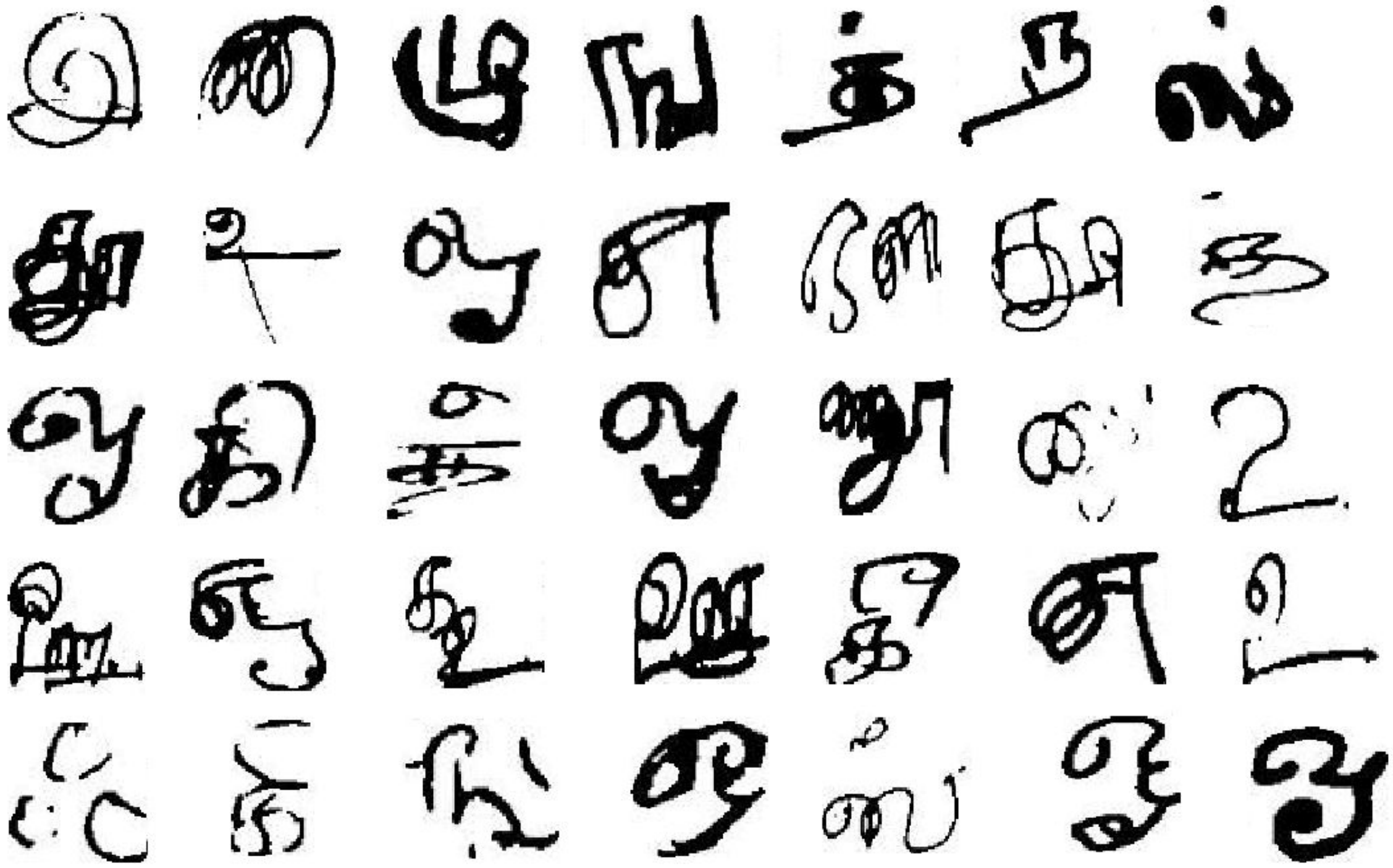

•

Fig.7(a) Sample Offline images in the database

characteristics of scanned handwritten samples such as slight distortion due to overwriting, variable thickness of strokes and stroke discontinuity which are typical in processing of scanned documents.

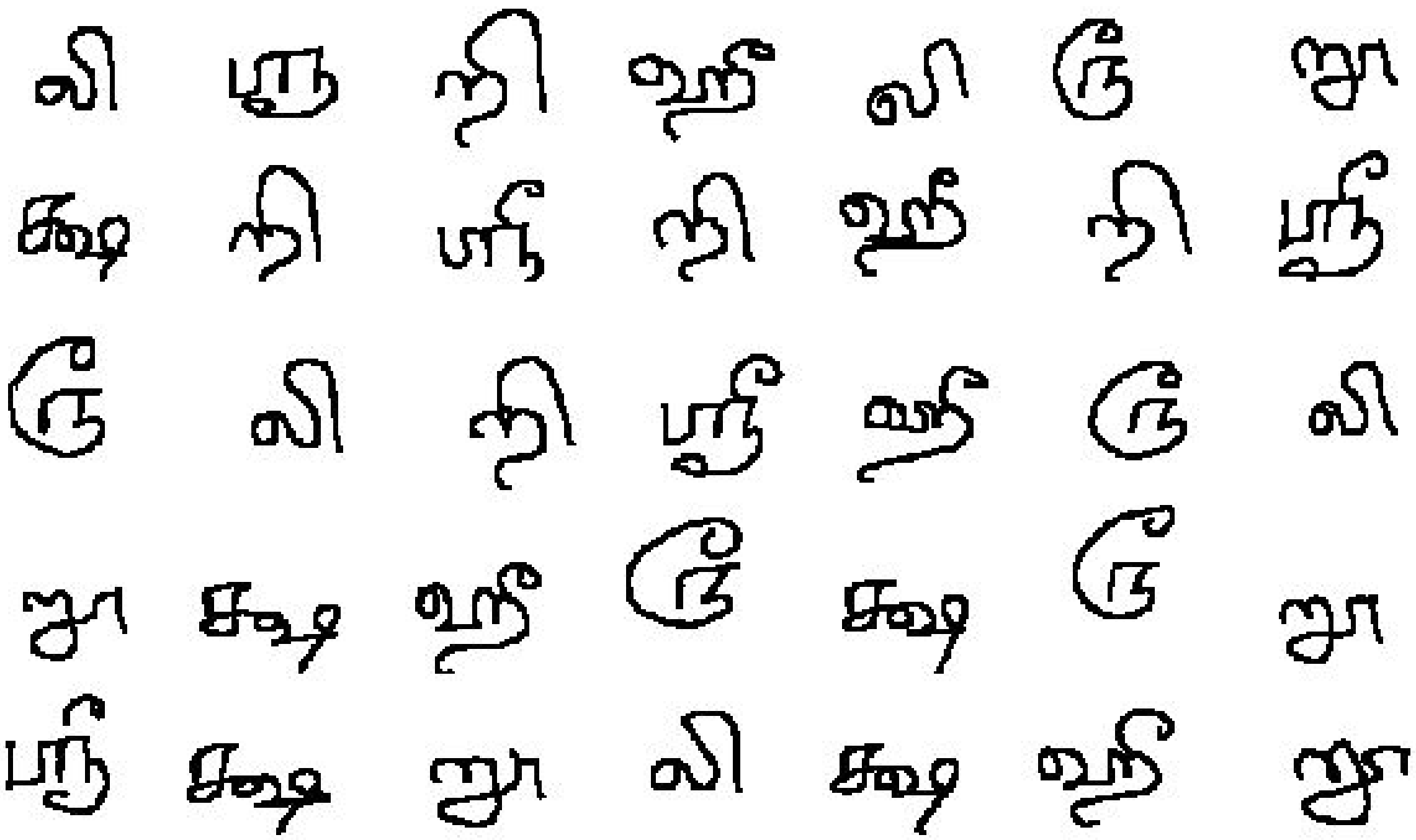

Fig.7(b) Sample Online images in the database

### 3.3: Sample Verification:

Two levels of verification was carried out to ensure that samples extracted are true representative of underlying Tamil character classes. Form-by-form verification takes place in Level-1 where as in Level-2 sample-by-sample verification happens. The complete verification process we followed is outlined in Fig.8.

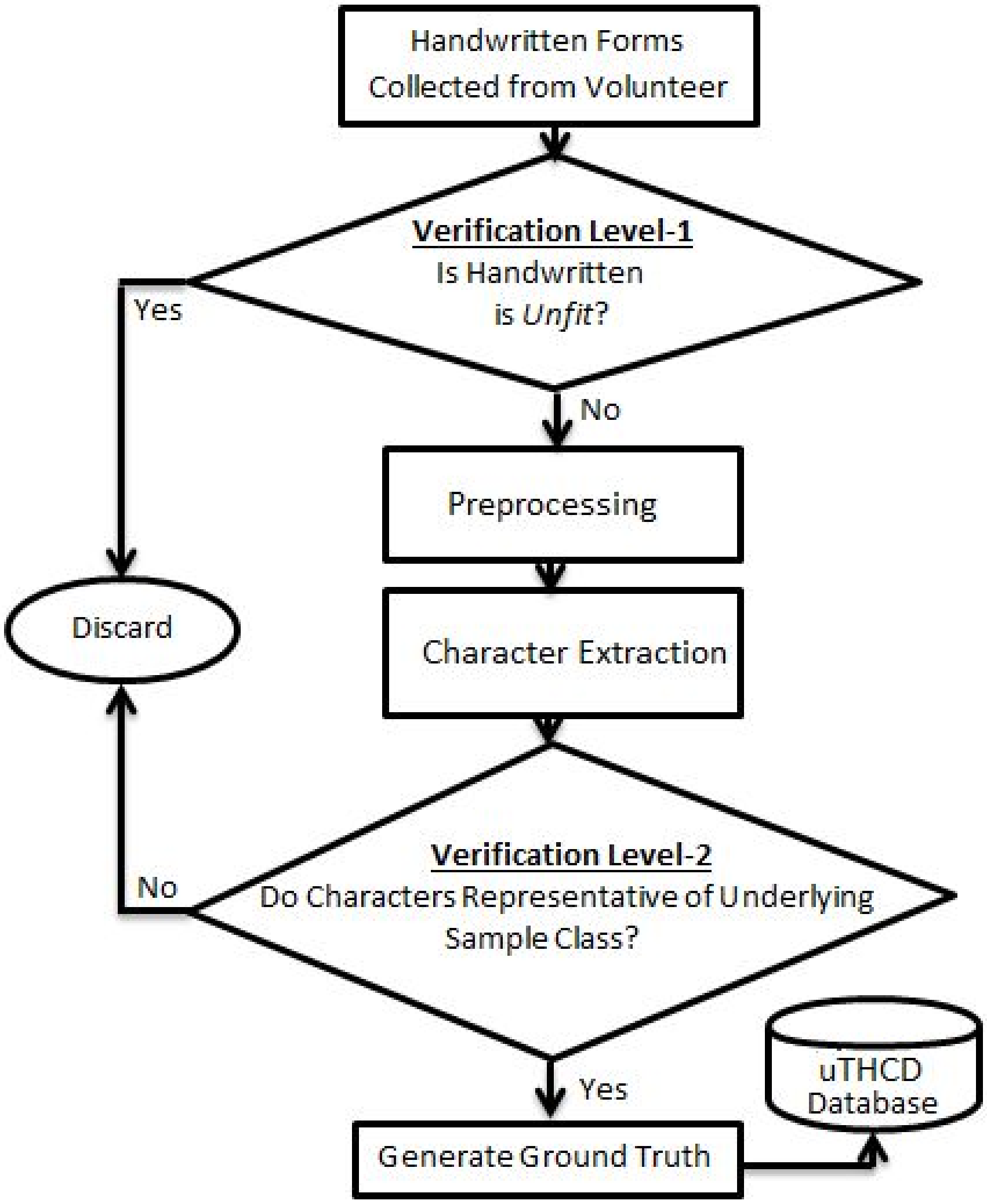

Fig.8 The Verification Process involved in uTHCD Database

In the first level of verification, we classify each scanned form or grid file (used in online data collection) into three categories:

1. Fit: If the writing style of 90% or more of the samples in the form represents the corresponding character classes.
2. Relatively Fit: If at least 50% of the samples in the form represent the corresponding character classes.
3. Unfit: If there are a considerable percentage of samples in the form that deviate much from the representative style of the character.

The data deemed as unfit are excluded from the rest of the process and it will not be collected in the eventual database. The remaining data in the fit and relatively fit categories are subject to preprocessing and sample extraction process.

The second level of verification has two main objectives:

a) The sample extracted represents one of the 156 classes of Tamil script.
b) Ensure GT associated with the sample conforms to the underlying class.

This sample-by-sample manual verification takes place after the data file is preprocessed and samples are extracted from the *fit* and *relatively fit* categories. The samples will be either discarded (if sample doesn't represent any Tamil character class) or collected in appropriate class. Few important comments from level-II verification process are:

- Some samples had very bad writing styles that were even hard to be recognized correctly by humans. Such samples were eventually discarded.
- Many characters in Tamil script look exactly the same except for the presence or absence of a tiny loop, dot/hole component etc. This minor inter-class variation sometimes causes a character written in one class to be considered in an appropriate class that resembles best.
- Some unintended scratch marks by volunteers accompanied the samples. However, we retained samples by only carefully removing such marks instead of completely ignoring the whole sample. This was found only in samples collected in offline mode.

Finally for GT verification, we made use of the file naming convention (xxxxs_yyy) we followed during the data extraction process. While doing sample-by-sample verification we also cross verified the GT (ie yyy) from the filename to ensure it indeed belongs to the right class. In this way, a simple python script later on helped us to generate a GT file with two columns of data that has filenames and corresponding class number as the ground truth.

### 3.4: The uTHCD: unconstrained Tamil Handwritten Character Database

The sample thus collected after arduous processes is our current uTHDC database with provision for subsequent expansion. Each sample will have a width and height ranging from 100-150 pixels which means that the characters are not centered. There were varying numbers of samples in each class but for the sake of consistency and ease of future expansion we limited the number of samples in each class to be approximately 600. All 156 character classes of this database are as shown in Fig.9**.** The mapping of unicode sequences with 156 classes in the uTHDC is given in the appendix A1.

The database has been randomly split into train and test sets respectively with 70% and 30% of total samples. The data has been distributed uniformly as both offline and online samples are included in the train and test folders.. Each set has a separate GT file. The database will be made publicly available for download in bi-level TIFF format as well as Hierarchical Data Format(HDF) to facilitate quick benchmarking. The details of data arrangement inside a HDF5 file (.h5 extension) along with supporting python code to extract the train and test set is given in Appendix A2.

| | | | | | | | | | | | | | | | |
|---|---|---|---|---|---|---|---|---|---|---|---|---|---|---|---|
| 0 | ◌ா | 20 | ச் | 40 | டி | 60 | நு | 80 | ர் | 100 | றி | 120 | னு | 140 | ஸீ |
| 1 | அ | 21 | ச | 41 | டீ | 61 | நூ | 81 | ர | 101 | றீ | 121 | ஷி | 141 | ஸு |
| 2 | ஆ | 22 | சி | 42 | டு | 62 | ப் | 82 | ரி | 102 | று | 122 | ஷீ | 142 | ஸூ |
| 3 | இ | 23 | சீ | 43 | டூ | 63 | ப | 83 | ரீ | 103 | றூ | 123 | ஷு | 143 | ஷ |
| 4 | ஈ | 24 | சு | 44 | ண் | 64 | பி | 84 | ரு | 104 | வ் | 124 | ஷூ | 144 | ஷ் |
| 5 | உ | 25 | சூ | 45 | ண | 65 | பீ | 85 | ரூ | 105 | வ | 125 | க்ஷ | 145 | னூ |
| 6 | ஊ | 26 | ங் | 46 | ணி | 66 | பு | 86 | ல் | 106 | வி | 126 | க்ஷ் | 146 | ஸ்ரீ |
| 7 | எ | 27 | ங | 47 | ணீ | 67 | பூ | 87 | ல | 107 | வீ | 127 | க்ஷி | 147 | க்ஷூ |
| 8 | ஏ | 28 | ஙி | 48 | ணு | 68 | ம் | 88 | லி | 108 | வு | 128 | க்ஷீ | 148 | ஜ |
| 9 | ஐ | 29 | ஙீ | 49 | ணூ | 69 | ம | 89 | லீ | 109 | வூ | 129 | ஜு | 149 | ஜ் |
| 10 | ஒ | 30 | ஙு | 50 | த் | 70 | மி | 90 | லு | 110 | ழ் | 130 | ஜூ | 150 | ஜி |
| 11 | ஓ | 31 | ஙூ | 51 | த | 71 | மீ | 91 | லூ | 111 | ழ | 131 | ஹ | 151 | ஜீ |
| 12 | ஔ | 32 | ஞ் | 52 | தி | 72 | மு | 92 | ள் | 112 | ழி | 132 | ஹ் | 152 | க்ஷு |
| 13 | ◌ஃ | 33 | ஞ | 53 | தீ | 73 | மூ | 93 | ள | 113 | ழீ | 133 | ஹி | 153 | ெ◌ |
| 14 | க் | 34 | ஞி | 54 | து | 74 | ய் | 94 | ளி | 114 | ழு | 134 | ஹீ | 154 | ே◌ |
| 15 | க | 35 | ஞீ | 55 | தூ | 75 | ய | 95 | ளீ | 115 | ழூ | 135 | ஹு | 155 | ை◌ |
| 16 | கி | 36 | ஞு | 56 | ந் | 76 | யி | 96 | ளு | 116 | ன் | 136 | ஹூ | | |
| 17 | கீ | 37 | ஞூ | 57 | ந | 77 | யீ | 97 | ளூ | 117 | ன | 137 | ஸ | | |
| 18 | கு | 38 | ட் | 58 | நி | 78 | யு | 98 | ற் | 118 | னி | 138 | ஸ் | | |
| 19 | கூ | 39 | ட | 59 | நீ | 79 | யூ | 99 | ற | 119 | னீ | 139 | ஸி | | |

Fig.9 Mapping of Class Number and Tamil Characters

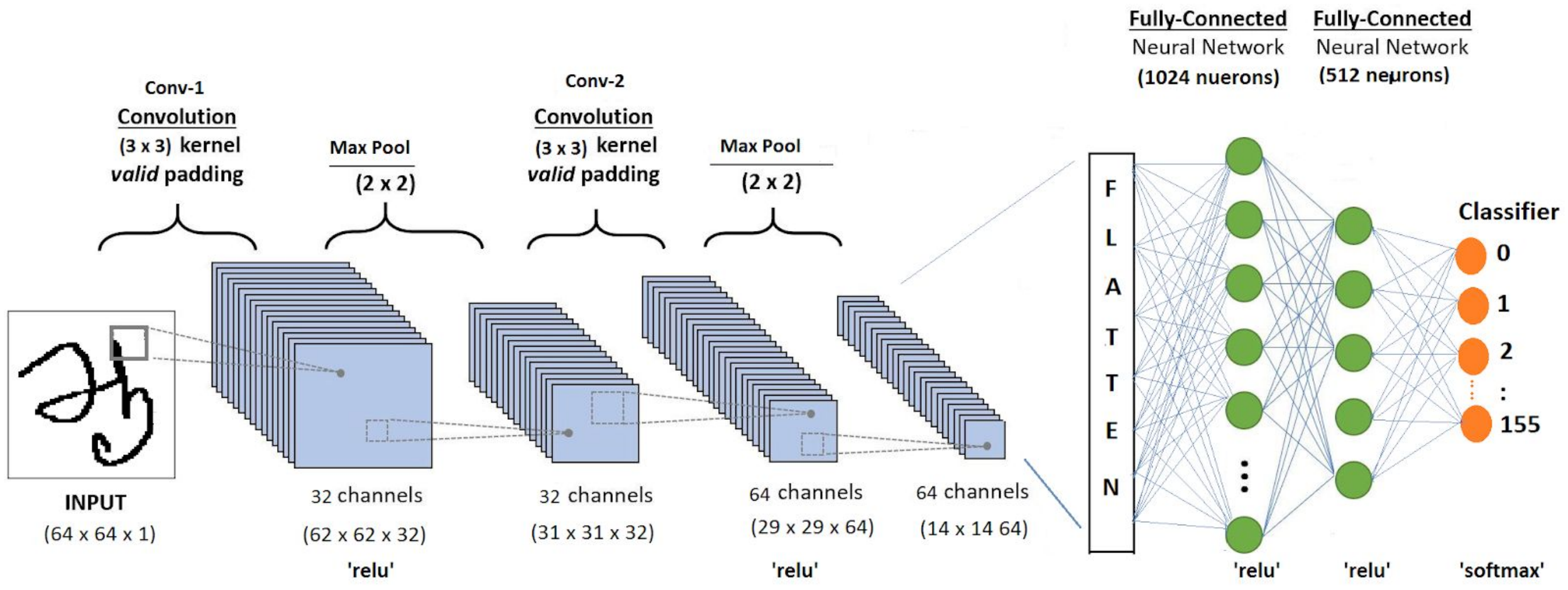

**Fig. 10 A Convolutional Neural Network Architecture**

## 4. Benchmarking:

Convolutional Neural Networks (CNNs) have dominated most of the recent technological advancements in the field of Artificial Intelligence especially in image recognition tasks. The main reason for that is the availability of large dataset, massively parallelizable GPUs and a wide spectrum of open deep learning frameworks.In this section, we present an ideal experimental set up for the proposed uTHCD dataset for the recognition of handwritten Tamil characters using the CNNs.

### 4.1: Brief about CNN:

CNNs are vastly popular for its ability to circumvent the high dimensionality problem and at the same time extracting robust features for computer vision tasks. There are three main building blocks of a CNN: i) convolutional layers, ii) zero or more pooling layers and iii) one or more fully connected layers. A variety of different architectures have been proposed in the literature by varying these building blocks [30, 31, 32, 33] which have obtained exceptionally good results.

4.1.1: Convolutional Layer: In a convolutional layer, an *mxn* image will convolve with a *fxf* filter ***k*** (or kernell) resulting in a volume ***C*** whose dimension depends on the number of filters, amount of striding ***s***, and whether or not padding ***p*** is applied. Activation function is applied in this layer after adding a bias to the resultant volume. The feature maps resulting from this process are robust and as the process gets deeper in the network certain higher level features are well captured. The number of trainable parameters in CNNs are comparatively very low compared to traditional fully connected ones.

4.1.2: Pooling Layer: The pooling layer usually follows the convolutional layer and is primarily used to scale down the number of features. The output of this layer is not trainable and hence no activation function is performed in this layer. There are different types of pooling algorithms available such as max pooling, average pooling etc. Suppose if C is of size mxn, a window of size pxp is moved all over the image with stride s and usually without padding. The max pooling algorithm for instance will record only the maximum value in the pxp region of volume thus resulting in the much reduced volume for subsequent layers.

4.1.3: Fully Connected Layer: Finally, the resulting volume from the preceding layer will be flattened out in the form of a vector and fed to a fully connected layer. The neurons in this layer will connect to all features resulting from the previous layer and hence the number of parameters are usually very high compared to convolutional layers. The output layer succeeds the fully connected layers usually with a softmax activation function.

**4.2: The CNN Architecture:**

The CNN architecture we have used for this as shown in the Fig. 10. It has 2 convolutional layers, 2 pooling layers and 2 fully connected layers. The input layer contains 64x64 sized images. The architecture can be represented as: 32C3-MP2-64C3-MP2-1024N-5128N, where nCj indicates a convolutional layer with n filters and jxj kernel, MPk refers to a max pool layers with kxk kernel and fN refers to a fully connected layer with N neurons. The output layer is a classifier with usual softmax activation function to classify 156 classes.

The CNN architecture we have demonstrated here is to establish a baseline accuracy for our experimentation with uTHCD. Apart from the architecture, there are several hyperparameters such as batch size, epochs, learning rate, activation functions etc that affects the overall accuracy of the model. In order to establish the baseline accuracy, we have used values for hyperparameters which are generally used in literature. For the rest of the experiments, unless explicitly mentioned, the hyperparameter values used are shown in the table 3.

**Table 3. Hyperparameters values used in the CNN Architecture**

| **Hyperparameters** | **Values** |
|---|---|
| Epochs | 200 |
| Learning Rate | 0.01 |
| Activation Function | Rectified Linear Unit (ReLU) |
| Optimizer | Adam |
| Batch size | 32 |
| Kernel Initializer | Glorot Uniform (Xavier) |

### 4.3 Training and Testing Process

As previously mentioned, the uTHCD database is split into 70:30 train and test sets respectively. The database contains a total of 90950 images in both train and test folders. We randomly segregated 62870 images into the train set and remaining images into the test set. This will be made public available to all researchers in both TIFF and HDF5 formats along with ground truth files. Out of 62870 images in the train set, the last 7870 images will be used as validation set during the run time. Each character image is cropped and resized to 64x64 pixels. The train-test-validation data split-up is shown in the Table-4.

**Table 4: Train-Test-Validation Split**

| Repository | Size |
|---|---|
| Train | 55000 |
| Test | 28080 |
| Validation | 7870 |

### 4.5 Results and Discussion

The baseline model used in this section resulted in a training accuracy of 99.61% with a validation accuracy of 92%. The plot of training and validation accuracy and loss are as demonstrated in Fig.11.

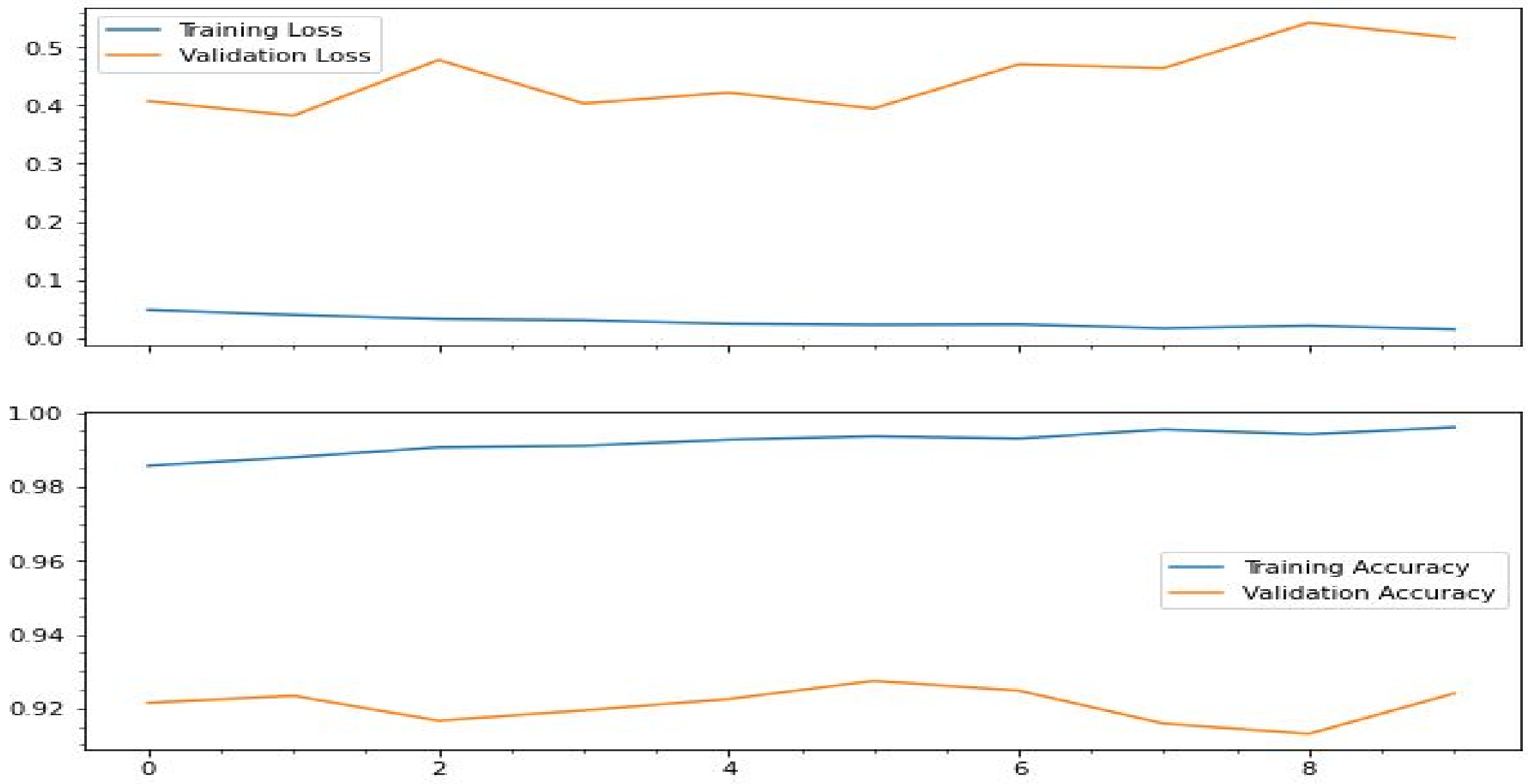

**Fig.11 Performance on Training vs Validation Sets.**

Next, we used the test set to evaluate the efficacy of the model. The test set obtained an accuracy of 87.67%. Considering the fact that human level performance can get nearly 0% error (which is also known as Bayes error/optimal error), these results indicate that the proposed model suffers from the high variance problem. This is because the proposed CNN model overfits the training data but performs significantly less with test data. This can be addressed in different ways such as by implementing regularization techniques, or even implementing a more appropriate network architecture. Ideally, the result should indicate a low variance and low bias situation where there is no much gap in the train and test error.

Based on the ground truth data, some example misclassified samples are shown in Fig.12. As can be seen from the figure that the most of the misclassifications are due to the low inter-class variability in the Tamil script. Fig. 13 shows the plot of misclassification errors for 156 classes. The top five misclassification errors are from classes தி, ஞு, ளு, மு and க்ஷு.

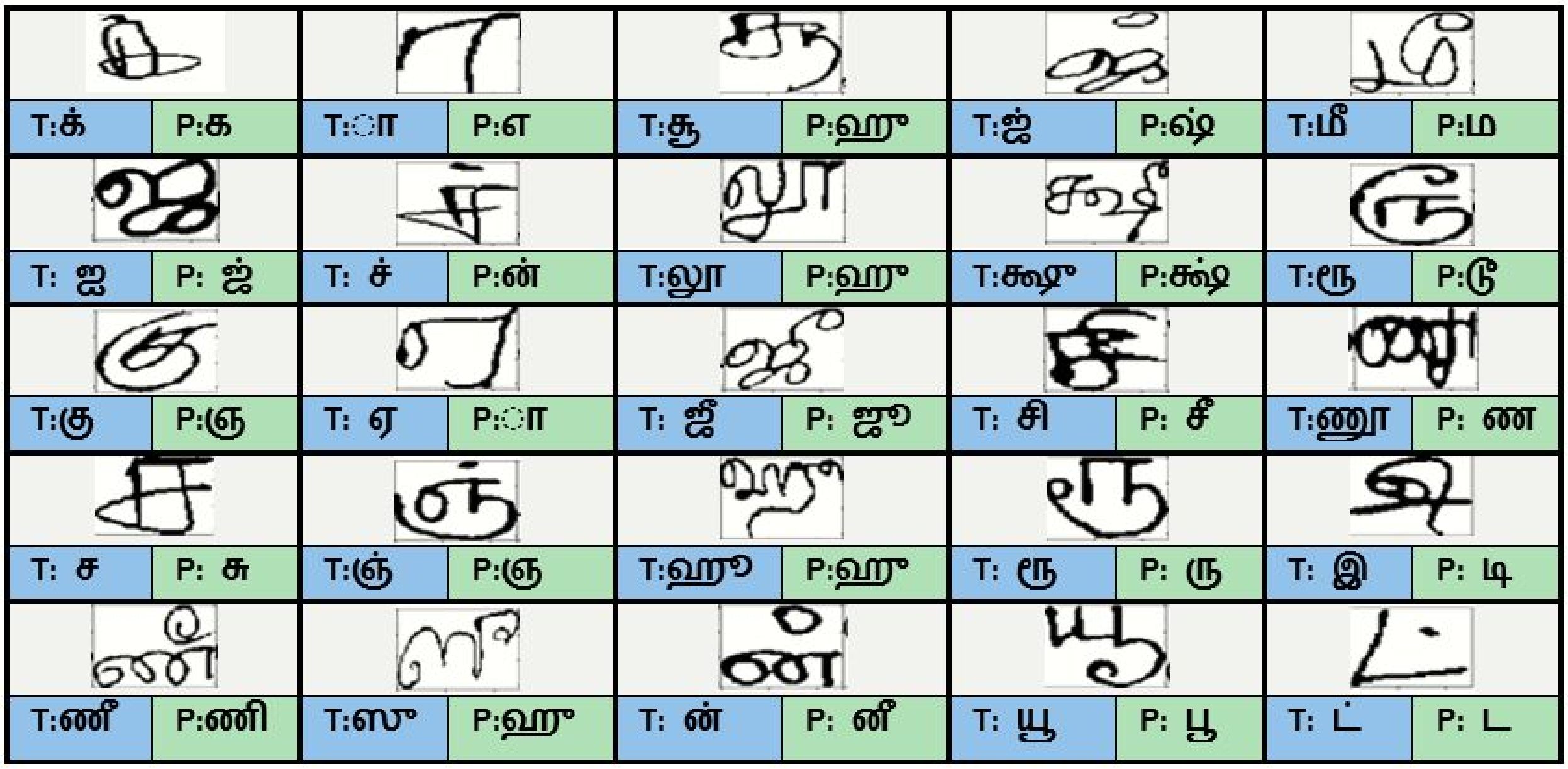

**Fig.12 Some Example Misclassified Samples from the Test Set (T:True, P:Predicted)**

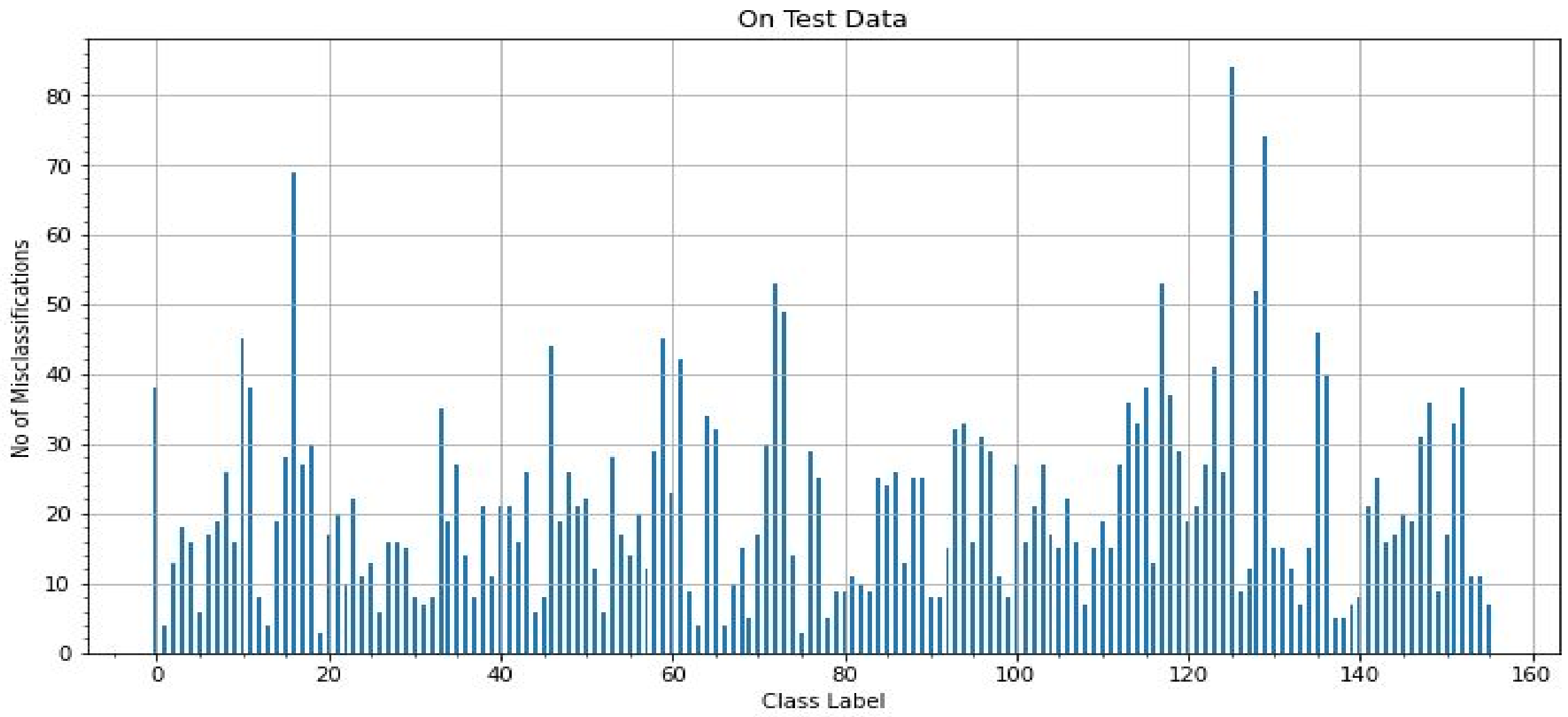

**Fig.13 Misclassification Count for 156 classes.**

The study of optimal values for hyperparameters of CNN is a challenging task and it is data dependent. In this section, we also present a series of experiments we carried out in order to tune these hyperparameters for optimal results using the uTHCD train-test-validation split as shown in Table-4.

Firstly, we conducted an experiment to determine the optimal value for batch size as it is a consequential hyperparameter in any network that controls the overall dynamics of the network architecture [36]. We varied the batch size starting from $2^5$ to $2^{12}$ insteps of a power of 2. The plot of train, test and validation accuracy vs batch size is shown in Fig.14. This was implemented with an *early stopping* callback observed on validation accuracy.

From figure 14, we can see that batch size is not having a significant impact on the accuracy. Hence, we can conclude that algorithms using this database on a CNN can prefer a minibatch gradient descent with 32 as an optimal value for the batch size that yields the same accuracy without compromising much on the computational burden and speed of the learning model.

Secondly, we conducted experiments to resolve the problem of high variance. One of the first things to try when there is a high variance problem is regularization. Dropout is a popular regularization technique that we used to resolve the high-variance problem our baseline model suffered. Hinton et al [34] who introduced the original dropout layer concept have used dropout (with probability=0.5) in each of the fully connected layers. It was not used on the convolutional layers. So, we conducted the first series of experiments by varying the dropout rate in the fully connected dense layers only. The result is shown in Fig.15. The dropout probability of 0.45 and 0.5 curtailed the gap between train and test accuracy thereby slight improvement towards high variance problem. It can be ascertained that these results comply with what was suggested in [34].

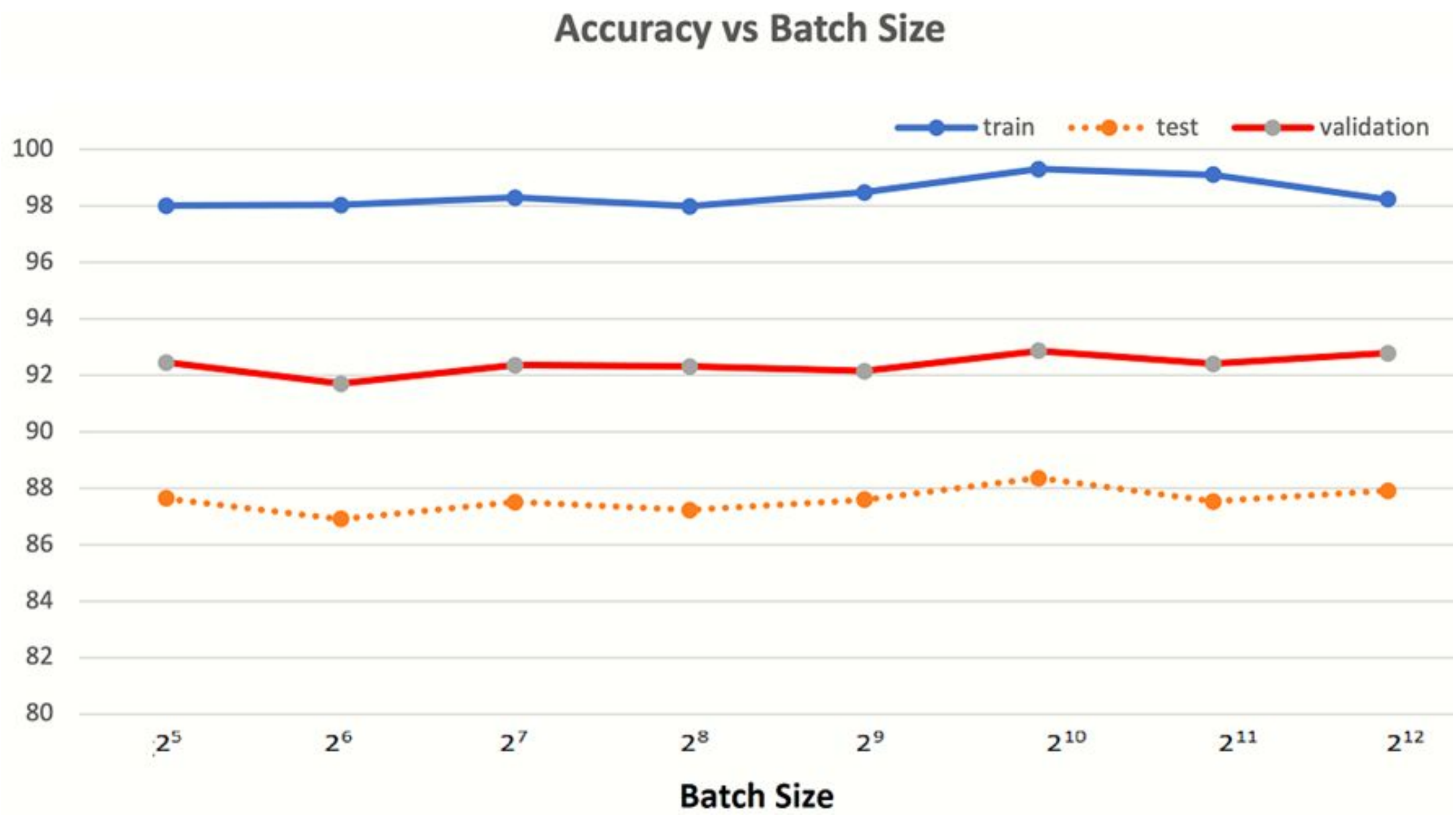

**Fig.14 Accuracy for varying batch sizes**

Of late a study proposed [35] applying the dropout in the convolutional layers as well, although with a much lower value of probability (0.1 or 0.2). Hence, to further mitigate the high variance problem, keeping the dropout rate of 0.5 in the dense layers, we conducted the next set of experiments by trying the dropout in the convolutional layers. The results are tabulated in table 5. The best accuracy on the test set was obtained for a dropout probability of 0.10 and 0.05 in conv1 and conv2 layers respectively.

**Table 5: Dropout in the convolutional layers**

| Conv1 | Conv2 | Train Acc | Test Acc | Validation Acc |
|---|---|---|---|---|
| 0.05 | 0.05 | 96.09 | 89.26 | 92.81 |
| 0.05 | 0.10 | 95.55 | 90.68 | 94.14 |
| 0.10 | 0.05 | 96.51 | **91.10** | 94.46 |
| 0.10 | 0.10 | 94.38 | 90.27 | 93.62 |

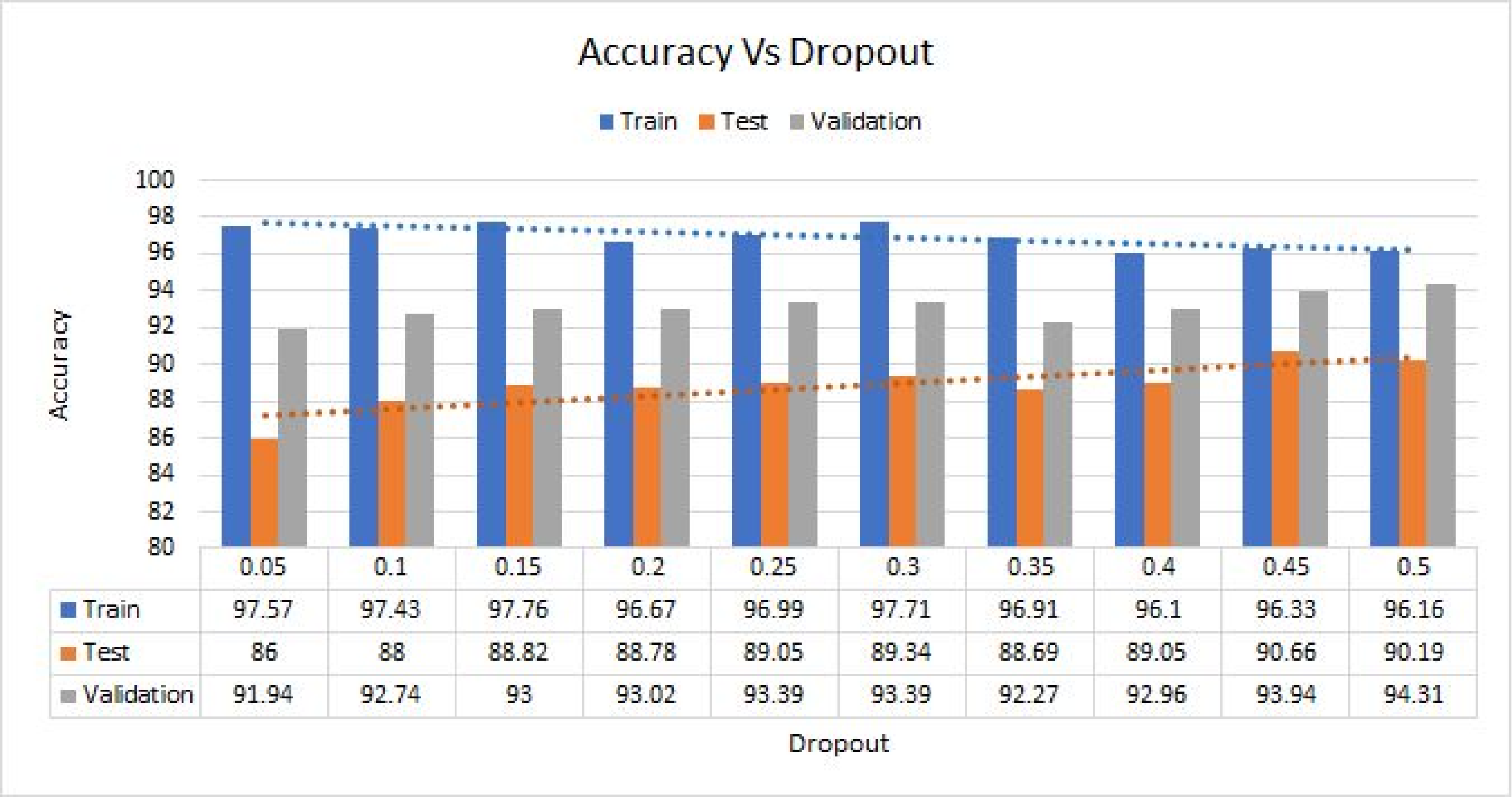

**Fig. 15 Effect of dropout regularization on train-test-validation accuracy**

There are several other hyperparameters such as number of hidden layers, no of units in each layer (fully connected), no of filters etc which deserve further tuning for optimal results.

## 5. Conclusion and Future Work

We embarked on this exhaustive work of database creation with an intention to enable the research community to propose robust algorithms, strengthen benchmarking in the field of Tamil OCR, and cater to ever evolving field of machine learning by the way of providing large collection of samples.

The samples were collected from a total of around 1000 volunteers to collect both offline and online samples. The volunteers of offline and online groups are not necessarily overlapping as this process happened during different time periods.The offline samples constitute around 30% of total samples and the rest of all samples are online samples. The final database contains a total of 90950 images. The samples are randomly split into train and test categories with a 70:30 proportion. We ensured train and test folders to contain both offline and online samples. The train and test folders contain a total of 62870 and 28080 samples respectively. Furthermore, the validation set can be formed out of the training set by slicing the last 7870 samples from the train set.

The database we collected will be made publicly available in two formats:

a) HDF5: The train and test set will be stored in python numpy multidimensional arrays by name x_train, y_train, x_test, y_test. The HDF5 file will be made available for download.

Availability of the database through this format will enable researchers to rapidly acquire the data which will be suitable for direct experimentation without much of preprocessing.

b) Raw Images: The total 90950 images will be made available as a compressed file format along ground truth file for both train and test folders. Researchers can then acquire this data according to their implementation preferences. The images will be in bi-level TIFF format.

The experiment we demonstrated for the CNN algorithm had a total of 55000, 28080 and 7870 in the train, test and validation sets respectively. An accuracy of 87.67% was achieved using the test dataset. However, the baseline model had the high-variance problem and we resolved this through an empirical study using the well known dropout regularization technique that resulted in train and test accuracy of 96.51 and 91.10 respectively.

In the immediate future, we would like to carry out some prominent works as mentioned below:

a) Augment this database by including more samples (especially offline samples) through the application of generative adversarial networks. This would enable the database to expand and pose more challenges to the ever evolving feature extraction algorithms and deep learning architectures. Data augmentation also helps in avoiding high variance problems that can affect the performance of a neural network architecture.
b) Provide a deeper CNN architecture that fits the data in a perfect manner taking care of high-bias and high variance problems. This study will also involve proposing optimal values for various hyperparameters.
c) To benchmark results with various feature extraction and classifier algorithms.

The creation of this database is expected to contribute to the efficient model development for Tamil OCR, facilitate the research community to objectively evaluate the state-of-the-algorithms and instigate more research work in the field of Tamil handwriting character recognition.

**Appendix A1. The Mapping of Class ID, Character and Unicode Sequence in the uTHDC.**

| Class # | Tamil Character | Unicode | Class # | Tamil Character | Unicode | Class # | Tamil Character | Unicode | Class # | Tamil Character | Unicode |
|---|---|---|---|---|---|---|---|---|---|---|---|
| 0 | ா | 0BBE | 29 | ஙீ | 0B99 0BC0 | 58 | நி | 0BA8 0BBF | 87 | ல | 0BB2 |
| 1 | அ | 0B85 | 30 | ஙு | 0B99 0BC1 | 59 | நீ | 0BA8 0BC0 | 88 | லி | 0BB2 0BBF |
| 2 | ஆ | 0B86 | 31 | ஙூ | 0B99 0BC2 | 60 | நு | 0BA8 0BC1 | 89 | லீ | 0BB2 0BC0 |
| 3 | இ | 0B87 | 32 | ஞ் | 0B9E 0BCD | 61 | நூ | 0BA8 0BC2 | 90 | லு | 0BB2 0BC1 |
| 4 | ஈ | 0B88 | 33 | ஞ | 0B9E | 62 | ப் | 0BAA 0BCD | 91 | லூ | 0BB2 0BC2 |
| 5 | உ | 0B89 | 34 | ஞி | 0B9E 0BBF | 63 | ப | 0BAA | 92 | ள் | 0BB3 0BCD |
| 6 | ஊ | 0B8A | 35 | ஞீ | 0B9E 0BC0 | 64 | பி | 0BAA 0BBF | 93 | ள | 0BB3 |
| 7 | எ | 0B8E | 36 | ஞு | 0B9E 0BC1 | 65 | பீ | 0BAA 0BC0 | 94 | ளி | 0BB3 0BBF |
| 8 | ஏ | 0B8F | 37 | ஞூ | 0B9E 0BC2 | 66 | பு | 0BAA 0BC1 | 95 | ளீ | 0BB3 0BC0 |
| 9 | ஐ | 0B90 | 38 | ட் | 0B9F 0BCD | 67 | பூ | 0BAA 0BC2 | 96 | ளு | 0BB3 0BC1 |
| 10 | ஒ | 0B92 | 39 | ட | 0B9F | 68 | ம் | 0BAE 0BCD | 97 | ளூ | 0BB3 0BC2 |
| 11 | ஓ | 0B93 | 40 | டி | 0B9F 0BBF | 69 | ம | 0BAE | 98 | ற் | 0BB1 0BCD |
| 12 | ஔ | 0B94 | 41 | டீ | 0B9F 0BC0 | 70 | மி | 0BAE 0BBF | 99 | ற | 0BB1 |
| 13 | ஃ | 0B83 | 42 | டு | 0B9F 0BC1 | 71 | மீ | 0BAE 0BC0 | 100 | றி | 0BB1 0BBF |
| 14 | க் | 0B95 0BCD | 43 | டூ | 0B9F 0BC2 | 72 | மு | 0BAE 0BC1 | 101 | றீ | 0BB1 0BC0 |
| 15 | க | 0B95 | 44 | ண் | 0BA3 0BCD | 73 | மூ | 0BAE 0BC2 | 102 | று | 0BB1 0BC1 |
| 16 | கி | 0B95 0BBF | 45 | ண | 0BA3 | 74 | ய் | 0BAF 0BCD | 103 | றூ | 0BB1 0BC2 |
| 17 | கீ | 0B95 0BC0 | 46 | ணி | 0BA3 0BBF | 75 | ய | 0BAF | 104 | வ் | 0BB5 0BCD |
| 18 | கு | 0B95 0BC1 | 47 | ணீ | 0BA3 0BC0 | 76 | யி | 0BAF 0BBF | 105 | வ | 0BB5 |
| 19 | கூ | 0B95 0BC2 | 48 | ணு | 0BA3 0BC1 | 77 | யீ | 0BAF 0BC0 | 106 | வி | 0BB5 0BBF |
| 20 | ச் | 0B9A 0BCD | 49 | ணூ | 0BA3 0BC2 | 78 | யு | 0BAF 0BC1 | 107 | வீ | 0BB5 0BC0 |
| 21 | ச | 0B9A | 50 | த் | 0BA4 0BCD | 79 | யூ | 0BAF 0BC2 | 108 | வு | 0BB5 0BC1 |
| 22 | சி | 0B9A 0BBF | 51 | த | 0BA4 | 80 | ர் | 0BB0 0BCD | 109 | வூ | 0BB5 0BC2 |
| 23 | சீ | 0B9A 0BC0 | 52 | தி | 0BA4 0BBF | 81 | ர | 0BB0 | 110 | ழ் | 0BB4 0BCD |
| 24 | சு | 0B9A 0BC1 | 53 | தீ | 0BA4 0BC0 | 82 | ரி | 0BB0 0BBF | 111 | ழ | 0BB4 |
| 25 | சூ | 0B9A 0BC2 | 54 | து | 0BA4 0BC1 | 83 | ரீ | 0BB0 0BC0 | 112 | ழி | 0BB4 0BBF |
| 26 | ங் | 0B99 0BCD | 55 | தூ | 0BA4 0BC2 | 84 | ரு | 0BB0 0BC1 | 113 | ழீ | 0BB4 0BC0 |
| 27 | ங | 0B99 | 56 | ந் | 0BA8 0BCD | 85 | ரூ | 0BB0 0BC2 | 114 | ழு | 0BB4 0BC1 |
| 28 | ஙி | 0B99 0BBF | 57 | ந | 0BA8 | 86 | ல் | 0BB2 0BCD | 115 | ழூ | 0BB4 0BC2 |

| Class # | Tamil Character | Unicode | Class # | Tamil Character | Unicode |
|---|---|---|---|---|---|
| 116 | ன் | 0BA9 0BCD | 136 | ஹூ | 0BB9 0BC2 |
| 117 | ன | 0BA9 | 137 | ஸ | 0BB8 |
| 118 | னி | 0BA9 0BBF | 138 | ஸ் | 0BB8 0BCD |
| 119 | னீ | 0BA9 0BC0 | 139 | ஸி | 0BB8 0BBF |
| 120 | னு | 0BA9 0BC1 | 140 | ஸீ | 0BB8 0BC0 |
| 121 | ஷி | 0BB7 0BBF | 141 | ஸு | 0BB8 0BC1 |
| 122 | ஷீ | 0BB7 0BC0 | 142 | ஸூ | 0BB8 0BC2 |
| 123 | ஷு | 0BB7 0BC1 | 143 | ஷ | 0BB7 |
| 124 | ஷூ | 0BB7 0BC2 | 144 | ஷ் | 0BB7 0BCD |
| 125 | க்ஷ | 0B95 0BCD 0BB7 | 145 | னூ | 0BA9 0BC2 |
| 126 | க்ஷ் | 0B95 0BCD 0BB7 0BCD | 146 | ஸ்ரீ | 0BB8 0BCD 0BB0 0BC0 |
| 127 | க்ஷி | 0B95 0BCD 0BB7 0BBF | 147 | க்ஷூ | 0B95 0BCD 0BB7 0BC2 |
| 128 | க்ஷீ | 0B95 0BCD 0BB7 0BC0 | 148 | ஜ | 0B9C |
| 129 | ஜு | 0B9C 0BC1 | 149 | ஜ் | 0B9C 0BCD |
| 130 | ஜூ | 0B9C 0BC2 | 150 | ஜி | 0B9C 0BBF |
| 131 | ஹ | 0BB9 | 151 | ஜீ | 0B9C 0BC0 |
| 132 | ஹ் | 0BB9 0BCD | 152 | க்ஷு | 0B95 0BCD 0BB7 0BC1 |
| 133 | ஹி | 0BB9 0BBF | 153 | ெ | 0BC6 |
| 134 | ஹீ | 0BB9 0BC0 | 154 | ே | 0BC7 |
| 135 | ஹு | 0BB9 0BC1 | 155 | ை | 0BC8 |

## Appendix A2 - Hierarchical Data Format (HDF) For uTHDC

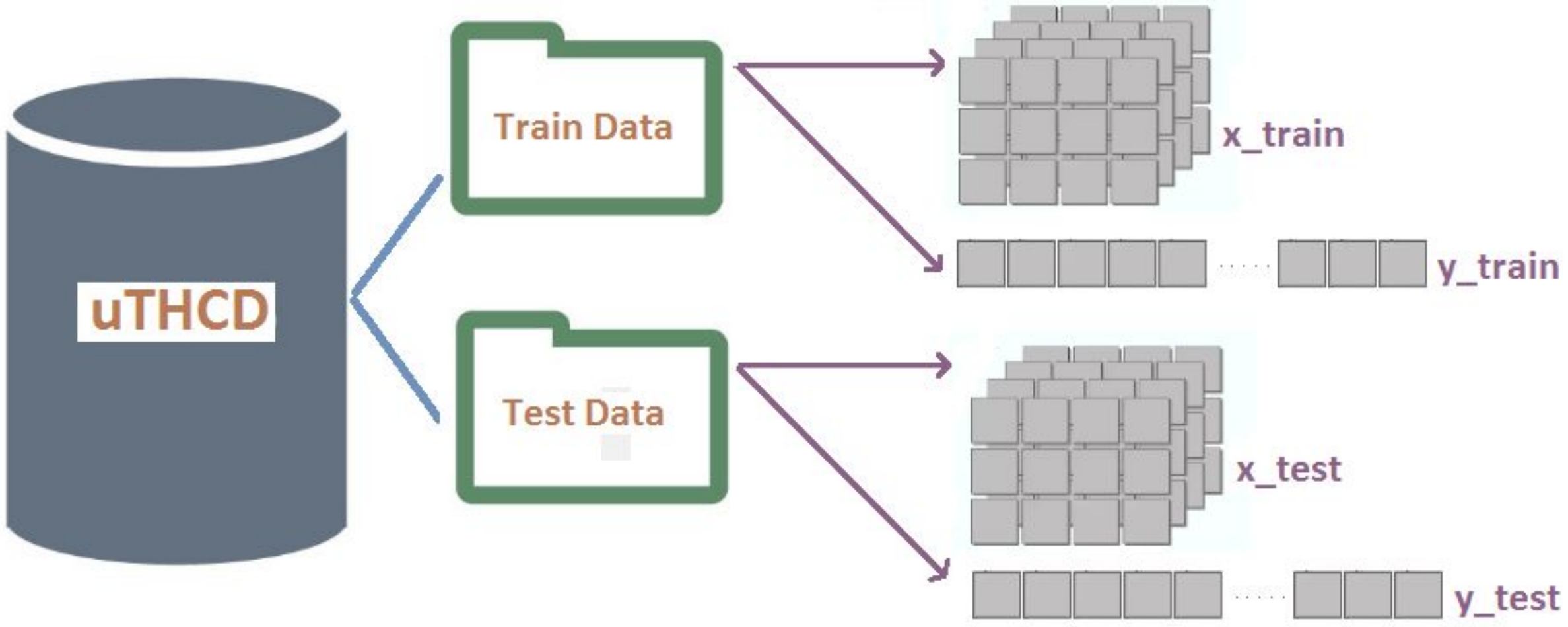

```
import numpy as np
import h5py

#%% READING FROM THE HDF5 File

with h5py.File('....../uTHCD_compressed.h5','r') as hdf:
    base_items = list(hdf.items())
    print('Items in the Base Directory:', base_items)

    G1 = hdf.get('Train Data')
    G1_items = list(G1.items())
    print('\n Items in Group 1:', G1_items)
      #Extract the train set.
    x_train = np.array(G1.get('x_train'))
    y_train = np.array(G1.get('y_train'))

    G2 = hdf.get('Test Data')
    G2_items = list(G2.items())
    print('Items in Group 2:', G2_items)
      #Extract the test set.
    x_test = np.array(G2.get('x_test'))
    y_test = np.array(G2.get('y_test'))
```

The HDF file uTHCD_compressed.h5 will be made publicly available. The data organization as shown in the pic. As illustrated in the code snippet above, the data necessary for experimentation (x_train, y_train, x_test, y_test) after all preprocessing (resizing to fixed 64x64, normalization) can be easily extracted for rapid experimentation purposes.